\documentclass[11pt,a4paper]{article}
\usepackage[hyperref]{acl2017}
\usepackage{times}
\usepackage{latexsym}

\usepackage{url}
\usepackage{flushend}
\usepackage{balance}
\usepackage{multirow}
\usepackage{graphicx}
\usepackage{comment}
\usepackage[utf8]{inputenc}
\usepackage{color,soul}
\usepackage{algorithm}
\usepackage{algpseudocode}

\newcommand{\ra}{\rightarrow}

\newcommand{\MC}{\multicolumn}
\newcommand{\MR}{\multirow}
\newcommand{\ds}{\displaystyle}

\aclfinalcopy 
\def\aclpaperid{49} 

\title{Learning Joint Multilingual Sentence Representations
	\\ with Neural Machine Translation \\
\vspace{44\baselineskip}
\textrm{\small Published at the 2nd ACL Workshop on Representation Learning for NLP, Vancouver, August 3rd 2017. \\[-4pt]
   This version contains slightly updated results and examples.}
\vspace{-43\baselineskip}
}

\author{Holger Schwenk  \\
  Facebook \\ AI Research \\
  {\tt schwenk@fb.com} \\
  \And
  Matthijs Douze \\
  Facebook \\ AI Research \\
  {\tt matthijs@fb.com}
 \\}

\date{}

\begin{document}
\maketitle
\begin{abstract}
In this paper, we use the framework of neural machine translation to learn joint sentence representations across six very different languages. Our aim is that a representation which is independent of the language, is likely to capture the underlying semantics.  We define a new cross-lingual similarity measure, compare up to 1.4M sentence representations and study the characteristics of close sentences.
We provide experimental evidence that sentences that are close in embedding space are indeed semantically highly related, but often have quite different structure and syntax.  These relations also hold when comparing sentences in different languages.

\end{abstract}

\section{Introduction}

It is today common practice to use distributed representations of words, often
called \textit{word embeddings}, in almost all NLP applications. It has been
shown that syntactic and semantic relations can be captured in this
embedding space, see for instance \cite{mikolov:2013:naacl}.
To process sequences of words, ie. sentences or small paragraphs, these
word embeddings need to be \textit{``combined''} into a representation of the
whole sequence. Common approaches include: simple techniques like bag-of-words
or some type of pooling, eg. \cite{Arora:2017:iclr_simple_sent_embed},
recursive neural networks, eg. \cite{Socher:2011_emnlp:recur_autoenc},
recurrent neural networks, in particular LSTMs, eg. \cite{cho:2014:emnlp_nmt},
convolutional neural networks, eg.
\cite{collobert:2008:icml_nlp,Zhang:2015_nips:text_convnet} or hierarchical
approaches, eg. \cite{Zhao:2015:arxiv_selfadapt_sentrep}.

In some NLP applications, both the input and output are sentences. A very
popular approach to handle such tasks is the  so-called
\textit{``encoder-decoder approach''}, also named
\textit{``sequence-to-sequence learning (seq2seq)''}.  The main idea is to
first encode the input sentence into an internal representation, and then to
generate the output sentence from this representation. A very successful
application of this paradigm is neural machine translation (NMT), see for
instance
\cite{Kalchbrenner:2013:emnlp_nmt,cho:2014:emnlp_nmt,Sutskever:2014:nips_nntrans}.
Current best practice is to use recurrent neural networks for the encoder and
decoder, but alternative architectures like convolutional networks have 
been also explored.

The performance of these vanilla seq2seq models substantially degrades with the
sequence length since it is difficult to encode long sequences into a single,
fixed-size representation.  A plausible solution is the so-called attention
mechanism \cite{Bahdanau:2015:iclr_nmt}: where the generation of each target
word is conditioned on a weighted subset of source words, instead of the full
sentence.  NMT has been also extended to handle several source and/or target
languages at once, with the goal of achieving better translation quality than
with separately trained NMT systems, in particular for under resourced
languages, see for instance
\cite{baidu:2015:acl_multinmt,zopf:2016:arxiv_multinmt,luong:2016:arxiv_multinmt,firat:2016:naacl_multinmt}.

In this work, we aim at learning \textit{multilingual sentence
representations}, i.e. which are independent of the language. Since we have to
compare these representations among each other, for the same or between
multiple languages, we only consider representations of fixed size.

There are many motivations to learn such a multilingual sentence representation, in particular:
\begin{itemize}
  \item 
  it is likely to capture the underlying semantics of the sentence
  (since the meaning is the only common characteristic of a sentence formulated in several languages);
  \item
  it has the potential to transfer many sentence processing applications to other languages (classification, sentiment analysis, semantic similarity, etc), without the need for language specific training data;
  \item
  it enables multilingual search;
  \item
  such representation could be considered as sort of a
  \textit{continuous space interlingua}.
\end{itemize}

To train these multilingual sentence embeddings we are using the framework of
NMT with multiple encoders and decoders. We first describe our model in detail,
relate it to existing research, and then present an experimental evaluation.

\section{Architecture}

\begin{figure}[b]
  \centering
  \includegraphics[width=0.5\textwidth]{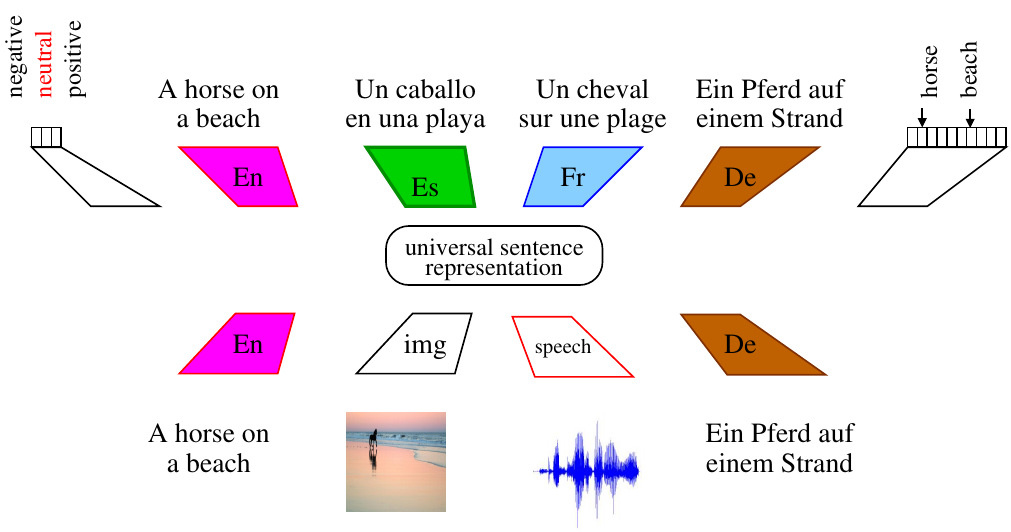}
  \caption{Generic multilingual and -modal encoder/decoder architecture.}
  \label{FigArchi}
\end{figure}

We propose to use multiple encoders and decoders, one for each source and
target language respectively.  The notion of multiple input languages can be
also extended to different modalities, e.g. speech and images. One can also
envision to add classification tasks, in addition to sequence generation.  Our
ultimate goal is to jointly train this generic architecture on many tasks at
once, to obtain a universal multilingual and -modal representation (see
illustration in Figure~\ref{FigArchi}).
To ease the comparison and search, we are focusing on representations of
fixed-size, independently of the length of the input (and output) sequence.
This choice has certainly an impact on the performance for very long sequences,
ie. in the order of more than fifty words, but we argue that such long
sentences are probably not very frequent in every day communication.  We would
also like to emphasize that the goal of this work is not to improve NMT (for
multiple languages), but to use the NMT framework to learn multilingual
sentence embeddings. Once the system is trained, the decoders are not used any
more. This means in particular that the usual attention mechanism cannot be used
since the attention weights are usually conditioned on the decoder outputs.  A
possible solution could be to condition the attention on the inputs only, for
instance so-called \textit{self-attention} \cite{Liu:2016:arxiv_innerattn} or
\textit{inner-attention} \cite{Lin:2017:iclr_struct_selfattn}.

To fix ideas, let us consider that we have corpora in $L$ different languages
which can be pairwise or $N$-way parallel, $N\le L$. This means that our
architecture is composed of $L$ encoders and $L$ decoders respectively.
However, this does not mean that we always provide input to all encoders, or
targets for all decoders, but we change the used models at each
mini-batch. One could for instance perform one mini-batch with two input
languages and one output language (which requires an 3-way parallel corpus),
and use one (different) input and output language in the next mini-batch (which
require a bitext). We call this \textit{partial training paths}. Note that
we can also use monolingual data in this framework, ie. the input and output
language is identical.

There are many possibilities to define partial training paths, with $1<M,N\le L$.
\begin{description}
  \item[\bf 1:1]
  translating from one source into one target language respectively.
  \item[\bf M:1]
  presenting simultaneously several source languages at the input.
  \item[\bf 1:N]
  translating from one source language into multiple target languages.
  \item[\bf M:N]
  this is a combination of the preceding two strategies and the most general approach. Remember that not all inputs and outputs need to be present at each training step.
\end{description}

\begin{figure*}[t]
  \centering
  \includegraphics[width=0.9\textwidth]{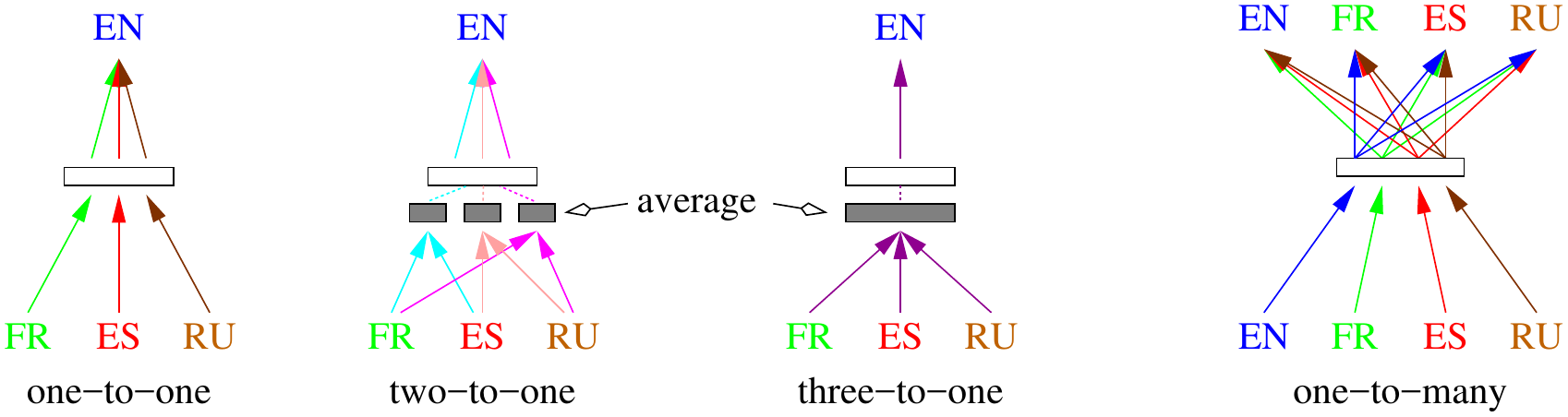}
  \caption[]{Possible partial training paths when four languages are available (En, Fr, Es and Ru).\\
    From left: \mbox{1:1}, \mbox{2:1} and \mbox{3:1} strategy, using En as common target language. \\
    Right: \mbox{1:3} strategy, translating from one source to the three other target languages.
  }
  \label{FigPaths}
\end{figure*}

Our goal is to learn joint sentence representations, which are as close as
possible when sentences are presented in different languages at the input.
If we use 1:1 training, changing the language pair at each mini-batch (input
and output), it is quite unlikely that the system would learn a common joint
representation which is independent of the source language. A variant of 1:1
training is to always use the same decoder, but many different encoders. Since
the decoder is shared for all the input languages, and the capacity of the
model is limited, there's an incentive for the system to use the same
representations for all the encoders.  This training strategy only requires
bitexts with one common language (usually English).  An important drawback,
however, is that we will not obtain an embedding of this common language since it
is never used at the input.\footnote{One could also use the common output
language at the input. This corresponds to training an auto-encoder which is
easier than a translation model and may have an negative impact.}

Using multiple languages at the input at the same time and combining the
corresponding sentence embeddings, ie. the \mbox{M:1} strategy, has in
principle the potential to learn joint sentence embeddings, if an appropriate
technique is used to combine the individual embeddings. The most
straightforward approach is to average the embeddings. This was used for
instance in \cite{firat:2016:emnlp_multinmt} in a multilingual NMT system with
attention.  The joint embedding could be also enforced by some type of
regularizer.  Again, having one dedicated output language makes it impossible
to learn a representation for it.

The \mbox{1:N} strategy is an interesting extension of~\mbox{1:1}. The idea is
translate from one input language simultaneously to all $L$-1 other languages,
excluding the one at the input (ie. no auto-encoder). The source and the set of
target languages is changed at each mini-batch. By these means, every input
language has at least one target language in common with all input
languages, and each target language has at least one input language in common.
On one hand hand, this strategy makes it possible to learn sentence embeddings for
all languages, but one the other hand, it requires $L$-way parallel training
data.  Although bitexts are usually used in MT, there are also several corpora
which can be aligned for more than two languages (eg. Eurpoarl, TED, UN).
Finally, the \mbox{N:M} strategy is the most generic one which combines all
above techniques.  These different training strategies are illustrated in
Figure~\ref{FigPaths} for four languages.

\subsection{Related work}

The use of multiple encoders and decoders was first studied in the context of
neural MT.  \citet{baidu:2015:acl_multinmt} used multiple decoders, i.e.
\mbox{1:N} training, to achieve improved NMT performance.
\citet{zopf:2016:arxiv_multinmt} and \citet{firat:2016:emnlp_multinmt}, on the
other hand, used multiple encoders, i.e. \mbox{M:1} training.  It's not
surprising that this complementarity improves MT quality, in comparison to one
input language only.  Many different configurations were explored by
\citep{luong:2016:arxiv_multinmt} for seq2seq models.
\citet{firat:2016:naacl_multinmt} were the first to use multiple encoders and
decoders with a shared attention mechanism.  This approach was further refined
to enable zero-resource NMT \cite{firat:2016:emnlp_multinmt}.  Alternatively,
it was proposed to handle multiple source and target languages with one encoder
and decoder only, using a special token to indicate the target language
\cite{google:2016:arxiv_multi_nmt} to enable zero-shot NMT.
To best of our knowledge, all these works focus on the improvement and
extensions of seq2seq modeling, and fixed-sized vector representations have not
analyzed in depth in a multilingual context.

Several publications consider joint representations in a multimodal context,
usually text and images, for instance
\cite{Frome:2013:nips_devise,Ngiam:2011:multmod_deep,Nakayama:2016:arxiv_encdec_pivot}.
The usual approach is to optimize a distance or correlation between the two
representations or \textit{predictive auto-encoders}
\cite{Chandar:2013:multlin_deep}.  The same approach was applied to
transliteration and captioning \cite{Saha:2016:arxiv_correncdec}.

There is a large body of research on sentence representations.  Common
approaches include: simple techniques like bag-of-words or some type of
pooling, eg \cite{Arora:2017:iclr_simple_sent_embed}, recursive NNs, eg.
\cite{Socher:2011_emnlp:recur_autoenc}, recurrent NNs, in particular LSTMs, eg.
\cite{cho:2014:emnlp_nmt}, convolutional NNs, eg.
\cite{collobert:2008:icml_nlp,Zhang:2015_nips:text_convnet} or hierarchical
approaches, eg. \cite{Zhao:2015:arxiv_selfadapt_sentrep}. In all these works,
the sentence representations are learned for one language only.  It is
important to note that our multiple encoder/decoder architecture and the
different training paths make no assumption on the type of encoder and decoder
used. In principle, all these sentence representations methods could be used.
This is left for future research.

There are several works on learning multilingual
representations at document level \citep{Hermann:2014:acl_multling,
Zhou:2016:acl_crossling, Pham:2015:multling}.
\citep{Hermann:2014:acl_multling} proposed a compositional vector model to
learn document level representations. Their model is based on bag of
words/bi-gram composition. \citep{Pham:2015:multling} directly learn a vector
representations for sentences in the absence of compositional property.
\citep{Zhou:2016:acl_crossling} learn bilingual document representation by
minimizing Euclidean distance between document representations and their
translation.

Other multilingual sentence representation learning techniques include BAE
\cite{Chandar:2013:multlin_deep} which trains bilingual autoencoders with the
objective of minimizing reconstruction error between two languages, and
BRAVE (Bilingual paRAgraph VEctors) \cite{Mougadala:2016:naacl_biword} which
learns both bilingual word embeddings and sentence embeddings from either
sentence-aligned parallel corpora (BRAVE-S), or label-aligned non-parallel
corpora (BRAVE-D).
%
%

Finally, many papers address the problem of learning bi- or multilingual word
representations which are used to perform cross-lingual document
classification. They are trained either on word alignments or
sentence-aligned parallel corpora, or both. I-Matrix
\cite{Klementiev:2012:coling_reuters} uses word alignments to do multi-task
learning, where each word is a single task and the objective is to move frequently
aligned words closer in the joint embeddings space. DWA (Distributed Word
Alignment) \cite{Kocisky:2014:acl_biword} learns word alignments and bilingual
word embeddings simultaneously using translation probability as objective.
Without using word alignments,  BilBOWA \cite{Gouws:2015:icml_bilbowa}
optimizes both monolingual and bilingual objectives, uses Skip-gram as
monolingual loss, while formulating the bilingual loss as Euclidean distance
between bag-of-words representations of aligned sentences. UnsupAlign
\cite{Luong:2015:aclwshop_biword} learns bilingual word embeddings by extending
the monolingual Skip-gram model with bilingual contexts based on word
alignments within the sentence. TransGram \cite{Coulmance:2015:emlp_crosslin}
is similar to \cite{Pham:2015:multling} but treats all words in the parallel
sentence as context words, thus eliminating the need for word alignments.

\section{Evaluation protocol}

An important question is how to evaluate multilingual joint sentence embeddings.
Let us first define some desired properties of such embeddings:
\begin{itemize}
  \item \textbf{multilingual closeness:}
  the representations of the same sentence for different languages should be as similar as possible;
  \item \textbf{semantic closeness:}
  similar sentences should be also close in the embeddings space, ie. sentences conveying the same meaning, but not necessarily the syntactic structure and word choice;
  \item \textbf{preservation of content:}
  sentence representations are usually used in the context of a task, eg. classification, multilingual NMT or semantic relatedness. This requires that enough information is preserved in the representations to perform the task;
  \item \textbf{scalability to many languages:}
  it is desirable that the metric can be extended to many languages without important computational cost or need for human labeling of data.
\end{itemize}

Two main approaches have been used in the literature to
evaluate multilingual sentence embeddings: 1) cross-lingual document
classification based on the Reuters corpus, first described in
\cite{Klementiev:2012:coling_reuters}; and 2) cross-lingual evaluation of
semantic textual similarity (in short STS). This task was first introduced in
the 2016 edition of SemEval \cite{Semeval:2016:sts_crossling}.
Both tasks focus on the evaluation of joint sentence representations of two
languages only. In the Reuters task, a document classifier is trained on
English sentence representations and then applied to texts in another language, and in the
opposite direction respectively.
STS seeks to measure the degree of semantic equivalence between two sentences
(or small paragraphs). Semantic similarity is expressed by a score between 0
(the two sentences are completely dissimilar) and 5 (the two sentences are
completely equivalent). In 2016, a cross lingual task was introduced (Es/En)
and extended to two more language pairs in 2017 (Ar/En and Tr/En).

In this work, we propose an additional evaluation framework for multilingual
joint representations, based on similarity search. Our metric can be
automatically calculated without the need of new human-labeled data and scaled
to many languages and large corpora.  We only need collections of $S$
sentences, and their translations in $L$ different languages, ie. $s_i^p,
i=1\ldots{}S, p=1\ldots{}L$. Such L-way parallel corpora are freely available,
for instance Europarl\footnote{\url{http://www.statmt.org/europarl/}} (20
languages), the UN corpus, 6 languages  \cite{Ziemski:2016:lrec_unv1}, or TED, 23
languages, \cite{cettolo:2012:eamt_ted}.

\algrenewcommand\algorithmicindent{0.8em}%
\begin{algorithm}
  \caption{Multilingual similarity search}
  \label{AlgoSim}
  \begin{algorithmic}[1]
  \State $L$: number of languages
  \State $S$: number of sentences
  \State $E_{pq}$: error between languages $p$ and $q$
  \State $R(s_i^p)$: embedding of a sentence
  \State $D()$: some distance metric
    \For{$p=1\ldots{}L$}
      \For{$q=1\ldots{}L, q\neq{}p$}
        \State $E_{pq} = 0$
        \For{$i=1\ldots{}S$}
          \If{$\ds \arg\min_{j=1\ldots{}S} D(R(s_i^p),R(s_j^q))\neq{}i$}
            \State $E_{pq}++$
          \EndIf
        \EndFor  
   	  \EndFor
    \EndFor
  \end{algorithmic}
\end{algorithm}

The details of our approach are given in algorithm~\ref{AlgoSim}. The basic
idea is to search the closest sentence in all $S$ sentences, and count an error
if it is not the reference translation. This requires the calculation of $S^2$
distance metrics and makes only sense when there are no duplicate sentences in
the corpus.  With increasing $S$ it may be also likely that the corpus contains
several alternative valid translations which could be closer than the reference
one. This is difficult to handle automatically at large scale and counted as
error by our algorithm.

Similarity search mainly evaluates the multilingual closeness property and can
be easily scaled to many languages. We will report results how the similarity
error rate is influenced by the number of language pairs and the size of the
corpus.
We have compared three distance metrics: L2, inner product and cosine.  In
general, cosine performed best. Note that all metrics are equivalent if the
vectors are normalized.

\section{Experimental evaluation}


We have performed all our experiments with the freely available UN corpus.
It contains about 12M sentences in six languages (En, Fr, Es, Ru, Ar and Zh).
We have used the version which is 6-way parallel (about 8.3M sentences).
This corpus comes with a predefined Dev and Test set (4000 sentences each).
We lowercase all texts, limit the length of the training data to 50 words and
use byte-pair encoding (BPE) with a 20k vocabulary. BPE allows to limit the
size of the decoder output vocabulary, it has only a small impact on the
sentence length ($\approx$ +20\%) and it showed similar or even superior
performance in NMT in comparison to many other techniques to limit the size of
the output vocabulary \citep{sennrich:2016:acl_bpe}. We have also found that
BPE is very robust to spelling errors which is important when handling
informal texts.

\subsection{Different network architectures}

\begin{table*}[t]
  \centering
  \begin{tabular}[t]{|l||*{4}{c|}}
  \hline
  \MR{2}{*}{System} & \MC{4}{c|}{Average Similarity Error} \\
         & efs & efsr & efsra & efsraz \\[-2pt]
  \MC{1}{|r||}{\#pairs:} & 6 &  10  &  15   &  21 \\
  \hline
  \hline
  \MC{5}{|l|}{\bf One-to-one systems:} \\
  efs-r   & 1.97\% &  -   &  -   &  - \\		
  efs-a   & 2.09\% &  -   &  -   &  - \\		
  efsr-a  & 1.90\% & 2.40\% &  -   &  - \\		
  efsra-z & 1.91\% & 2.26\% & 2.51\% &  - \\		
  \hline
  \MC{5}{|l|}{\bf One-to-many systems:} \\
  efsraz-all & 1.70\% & 1.97\% & 2.38\% & 2.59\% \\	
  \hline
  \MC{5}{|l|}{\bf One-to-many systems, nhid=1024:} \\
  efsraz-all & 1.36\% & 1.64\% & 1.89\% & 1.95\% \\	
  \hline
  \MC{5}{c}{} \\[-10pt]
  \MC{5}{c}{\bf Three layer LSTM, nhid=512} \\
  \MC{5}{c}{\bf Sentence representation: last LSTM state} \\
  \end{tabular}
  \hfill
  \begin{tabular}[t]{|l||*{4}{c|}}
  \hline
  \MR{2}{*}{System} & \MC{4}{c|}{Average Similarity Error} \\
         & efs & efsr & efsra & efsraz \\[-2pt]
  \MC{1}{|r||}{\#pairs:} & 6 &  10  &  15   &  21 \\
  \hline
  \hline
  \MC{5}{|l|}{\bf One-to-one systems:} \\
  efs-r   & 1.11\% &  -   &  -   &  - \\		
  efs-a   & 1.03\% &  -   &  -   &  - \\		
  efsr-a  & 1.11\% & 1.31\% &  -   &  - \\		
  efsra-z & 1.01\% & 1.19\% & 1.25\% & --  \\		
  \hline
  \MC{5}{|l|}{\bf One-to-many systems:} \\
  efsraz-all & 0.92\% & 1.07\% & 1.15\% & \bf 1.20\% \\	
  \hline
  \MC{5}{c}{} \\[18pt]
  \MC{5}{c}{\bf One layer BLSTM, nhid=512} \\
  \MC{5}{c}{\bf Sentence representation: max pooling} \\
  \end{tabular}
  \caption{Error rates of similarity search on the UN Dev corpus.
    Languages are abbreviated with the following letters:
    e=English, f=French, s=Spanish, r=Russian, a=Arabic, z=Chinese.
  }
  \label{TabSim}
\end{table*}

In this work we only consider stacked LSTMs as encoders and decoders. In the
vanilla seq2seq NMT model, the last state of the LSTM is used as sentence
representation.  There is also evidence that deeper architectures perform
better in NMT than shallow ones, eg.
\cite{baidu:2016:tacl_nmt,google:2016:arxiv_nmt}. Following this tendency, we
performed the first set of experiments with stacked LSTMs with three
512-dimensional hidden layers. Deeper architectures did not improve the
performance.

We then switched to using BLSTMs followed by max-pooling (element-wise over the
sequence length).  We are not aware of works which use max-pooling in an NMT
framework. One is indeed tempted to assume that max-pooling makes it more
difficult to create a target sentence which preserves all information from the
source sentence.  On the other hand, max-pooling is successfully used in
various sentence classification tasks,  eg. \cite{Alexis:2017}.  It should be
noted that the final sentence representation has twice the dimension of the
BLSTM hidden layer.

The word embeddings are of size 384 for all models.  We use vertical dropout
with a value of 0.2 and gradients are clipped at 2. The initial learning rate
is set to 0.01 and decreased each time performance on the Dev data
does not improve. Performance is measured by perplexity for the decoders and
similarity error at the embedding layer for the encoders. It is important to
note that the similarity error rate can be only calculated once the whole
development set is processed. Therefore it is not used to provide gradients to
the encoders.
Training is performed for up to five epochs with a batch size of 96. For the
smallest models, one iteration through the training data takes about 11h. Most
models converge after two to three epochs.

Table~\ref{TabSim} summarizes our results on the UN Dev corpus for several
systems using the one-to-one and one-to-many partial training paths. We compare
training of joint representations for three to six languages using LSTM or
BLSTM encoders.
In each column, we give the average similarity error over all $n(n+1)/2$
language pairs. As an example, the system trained with En, Fr, Es and Ru at the
input and Ar at the output (\textit{``efsr-a''} in the third line), achieves an
average similarity error of 1.90\% over 6 language pairs\footnote{En-Es, En-Fr,
Es-En, Es-Fr, Fr-En and Fr-Es.}, column \textit{``efs''}, and 2.40\% over 10
languages pairs\footnote{En-Es, En-Fr, En-Ru, Es-En, Es-Fr, Es-Ru, Fr-En,
Fr-Es, Fr-Ru, Ru-En, Ru-Es and Ru-Fr.}, column \textit{``efsr''}.

We can make the following observations.
First, using an BLSTM with max-pooling (Table~\ref{TabSim} right) performs much
better than an LSTM and using the last hidden state as sentence representation
(Table~\ref{TabSim} left). This was also observed for many monolingual tasks,
eg. \cite{Alexis:2017}.  This is particularly true when the number of languages
is increased.  This performance gain does not result from the increased
dimension of the sentence representation (2$\times$nhid) since an
1024-dimensional LSTM only achieves 1.36\% (see last line in Table~\ref{TabSim}
left).
Second, increasing the number of languages for which we seek a joint sentence
embedding does not seem to make the task harder: the system trained on all
languages achieves the same results (1.01\%) on three languages than when
training only on these languages (1.03\%).
Third, the one-to-many training strategy (\textit{efsraz-all}, 0.92\%) performs
better than \mbox{1:1} (\mbox{\it efsra-z}, 1.01\%).  In addition, it allows to
obtain a sentence embedding for all languages, while the common output language
is excluded in the \mbox{1:1} strategy.

Finally, we have explored whether deep architectures are needed when using an
BLSTM encoder and a max-pooling sentence representation (see
Table~\ref{TabSimBSLTM}).  We found no experimental evidence that stacking
several BLSTM layers is useful.

\begin{table*}[t]
  \centering
  \begin{tabular}[t]{|l||*{2}{c|}|*{3}{c|}*{3}{c|}}
  \hline
  \hline
  \MR{2}{*}{Network} & \MC{2}{c||}{LSTM + last} & \MC{6}{c|}{BLSTM + max-pooling} \\
                         & 3x512 & 3x1024 & 1x256 & 2x256 & 3x256 & 1x512 & 2x512 & 3x512 \\
  \hline
  \mbox{1:1}, efsra-z    &  2.51 &   --   &  1.44 &  1.21 & 1.52 &  1.32 &  1.25 & 1.41 \\
  \mbox{1:M}, efsraz-all &  2.38 &  1.89  &  1.27 &  1.30 & 1.27 & \bf 1.15 &  1.17 & 1.25 \\
  \hline
  \end{tabular}
  \caption{Error rates of similarity search on the UN Dev corpus
    for \textbf{five} language pairs (\textit{efsra}).
    Comparision of LSTMs and BLSTMs of different size and depth.
  }
  \label{TabSimBSLTM}
\end{table*}

\subsection{Many-to-one training strategies}

\begin{table}[b!]
  \centering
  \begin{tabular}[t]{|c||*{3}{c|}c|}
    \hline
    & \MC{3}{|c|}{\# input languages} & Similarity \\
    ID & 1 & 2 & 3 & Error \\
    \hline
    \hline
    \MC{5}{|l|}{\bf One M:1 strategy} \\
    1    &  1  &  -- &  -- & \bf 1.03\% \\
    2    &  -- & 0.5 &  -- & 1.85\% \\
    3    &  -- &  -- &   1 & 67.9\% \\
    \hline
    \MC{5}{|l|}{\bf Combining 1:1 and 2:1 strategies} \\
    $12_a$  & 0.9  & 0.05 & -- & 1.09\% \\
    $12_b$  & 0.8  & 0.10 & -- & 1.16\% \\
    $12_c$  & 0.7  & 0.15 & -- & 1.15\% \\
    $12_d$  & 0.6  & 0.20 & -- & 1.12\% \\ 
    $12_e$  & 0.5  & 0.25 & -- & 1.22\% \\
    \hline
    \MC{5}{|l|}{\bf Combining 1:1 and 3:1 strategies} \\
    13   & 0.5  &  --  & 0.5 & 1.31\% \\ 
    \hline
    \MC{5}{|l|}{\bf Combining 1:1, 2:1 and 3:1 strategies} \\
    $123_a$ & 0.33 & 0.16 & 0.33 & 1.32\% \\ 
    $123_b$ & 0.25 & 0.25 & 0.25 & 1.35\% \\ 
    \hline
  \end{tabular}
  \caption{Different \mbox{M:1} strategies for three input languages
	(system \textit{efs-a}).
	The baseline with the 1:1 strategy is 1.03\% (line with ID~1).
  }
  \label{TabSimMulti}
\end{table}

In this section, we study two \mbox{M:1} training strategies, namely \mbox{2:1}
and \mbox{3:1}.  Since the number of combinations quickly increases with the
number of input languages, we limit these experiences to three input languages
(system \textit{efs-a}). In that case, we have three \mbox{1:1} training paths
(En$\ra$Ar, Fr$\ra$Ar and Es$\ra$Ar), three \mbox{2:1} training paths
(\mbox{En+Fr$\ra$Ar}, \mbox{En+Es$\ra$Ar} and \mbox{Fr+Es$\ra$Ar}) and one
\mbox{3:1} configuration (\mbox{En+Fr+Es$\ra$Ar}).  This is illustrated in
Figure~\ref{FigPaths}.
To obtain efficient training, we use homogeneous mini-batches, ie. the number
of encoders and decoders is constant. Examples in a mini-batch are sampled
according to a coefficient.  In order to make a fair comparison, these
resampling coefficient were chosen so that each encoders always sees the same
number of sentences (roughly 8.3M).  We refer to the different runs with an ID
(first column in Table~\ref{TabSimMulti}).
As an example, for the experiment with ID~l2$_a$, 90\% of the mini-batches are
\mbox{1:1} and 5\% are \mbox{2:1}. Note that that the \mbox{2:1} samples have a
coefficient of 0.05 since two encoders are simultaneously used.

The first striking result is that presenting all input languages at once and
averaging the three sentence representations (\mbox{3:1}, ID 3) does not allow
to learn joint representations.
We are however able to learn joint representations with the \mbox{2:1} strategy
(ID~2), but the performance is worse than the \mbox{1:1} baseline (1.85\%
versus 1.03\%).
We are also tried to alternate between \mbox{1:1} and \mbox{2:1} mini-batches with
increasing resampling coefficients (ID 12$_a$ to 12$_e$). The idea is that each
encoder learns to provide a sentence representation when used alone and when
used with another one.  However, we observe that adding \mbox{2:1} training
paths is not useful: the similarity error increases.
The same observation holds when adding \mbox{3:1} training paths (ID~13 and
123).
Overall, we were not able to improve the baseline of 1.03\% similarity error
obtained with a simple \mbox{1:1} training strategy.  Therefore, we did not try
the even more complex \mbox{M:N} paths.  This failure could be attributed to
the fact that we simply average multiple sentence representations.  In future
research, we will investigate other possibilities, eg. based on correlation like
proposed in \cite{Saha:2016:arxiv_correncdec,Chandar:2016:nc_corrnn}.

Detailed similarity search error rates for all \textbf{six} languages,
including Zh, of our best system are given in Table~\ref{TabSimAll6}. Overall,
the error rates vary only slightly from the average of 1.2\% although the six
languages differ significantly with respect to morphology, inflection, word
order, etc. In particular, Chinese is handled as well as the other languages.
This is in nice contrast to many other NLP application, in particular NMT, for
which the performances on Chinese are significantly below those of other
languages. All error rates are below 1.7\%.

\begin{table}[t]
  \centering
  \begin{tabular}[t]{|@{\,\,}r@{\,\,}||*{6}{@{\,\,}c@{\,\,}|}|@{\,\,}c@{\,\,}|}
  \hline
      & \MC{7}{c|}{Target language} \\
  Src   & En & Fr & Es & Ru & Ar & Zh & All \\
    \hline
     En &  --  & 1.10 & 0.70 & 1.07 & 1.05 & 1.15 & 1.02 \\
     Fr & 0.97 &  --  & 0.95 & 1.55 & 1.65 & 1.68 & 1.36 \\
     Es & 0.68 & 1.10 &  --  & 1.20 & 1.35 & 1.27 & 1.12 \\
     Ru & 0.78 & 1.52 & 1.23 &  --  & 1.32 & 1.32 & 1.23 \\
     Ar & 0.78 & 1.52 & 1.07 & 1.48 &  --  & 1.23 & 1.22 \\
     Zh & 0.97 & 1.55 & 1.12 & 1.35 & 1.30 &  --  & 1.26 \\
    \hline
    All & 0.83 & 1.36 & 1.02 & 1.33 & 1.33 & 1.33 & \bf 1.20 \\
    \hline
  \end{tabular}
  \caption{Pair-wise error rates of similarity search for 6 languages (UN Dev).
    Training was performed with a one layer BLSTM with 512 hiddens, max-pooling
     and the \textit{``efsraz-all''} strategy.
  }
  \label{TabSimAll6}
\end{table}

\subsection{Large scale out-of domain similarity search}

In this section, we evaluate our sentence representation on out-of domain data.
We are not aware of another huge corpus which is 6-way parallel for the same
languages than the UN corpus. Therefore, we have selected the Europarl corpus
and limit our study to three common languages (En, Fr and Es).  After
excluding duplicates and limiting the sentence length to fifty tokens, we
dispose of almost 1.5 million 3-way parallel sentences. 

The two training strategies \mbox{\it ``efsra-z''} and \mbox{\it
``efsraz-all''} achieve the same similarity error rate of about 7.7\%.  We
argue that this is an interesting result given the size of the corpus (1.46M
sentences) and the fact that it contains several sentences which are very
similar (e.g. \textit{``The session resumes on DATE''}).  Using the last state
of an LSTM 3x512 achieves an error rate of 12.2\%.
Evaluating the similarity error requires the calculation of 1.46M$^2$ distances
for each language pair.  This can be very efficiently performed with the FAISS
open-source toolkit \cite{FAISS:2017:arxiv} which offers many options to
increase the speed of nearest neighbor search. Its implementation of
brute-force L2 search was sufficient for our purposes.

\subsection{Examples of multilingual search}

On the next page, we give several examples of similarity search. For each
example, we give the query and the five closest sentences. Remember that we
use the cosine distance, i.e. the value of 1.0 is a perfect match and smaller
values are worse.

The first example in Table~\ref{Ex1a} shows two simple query sentences for
which four paraphrases were found in the Europarl corpus. The value of the cosine
distance clearly indicates the closeness (the last three sentences in Table~\ref{Ex1a}
left only share some aspects).
A more complicated
query sentence is used in the second example (see Table~\ref{Ex1b}).  For such
longer sentences, it is unlikely to find several perfect paraphrases in the
indexed corpus.  However, the system was able to retrieve sentences which share
a lot of the meaning of the query: all cover the topic \textit{``punishment of
(sexual) crimes, independently of the country the crime is committed in''}.
Finally, examples of cross-lingual similarity search are given in
Tables~\ref{Ex2a} and \ref{Ex2b}.  In the first example, all five nearest
French and Spanish sentences have very similar cosine distances, and all 
are indeed semantically related.

Table~\ref{Ex2b} gives an example where not all retrieved sentences have
similar cosine distances.  The closest sentence is the correct translation, for
French and for Spanish.  Both second closest sentences are well related to the
query and also have a cosine distance close to the best scoring sentence.  The
third and following sentences are less related with the query, which is clearly
reflected in the substantially lower cosine distance.
It's interesting to note that the two closest sentences are all identical,
independently of the language.  This can be seen as experimental evidence of
the quality of the multilingual sentence embeddings.

\section{Conclusion}

We have shown that the framework of NMT with multiple encoders/decoders can be
used to learn joint fixed-size sentence representations which exhibit
interesting linguistic characteristics.  We have explored several training
paradigms which correspond to partial paths in the whole architecture.
We have proposed a new evaluation protocol of multilingual similarity search
which easily scales to many languages and large corpora.
We were able to obtain an average cross-lingual similarity error rate of 1.2\%
for all 21 languages pairs between six languages\footnote{English, French,
Spanish, Russian, Arabic and Chinese.} which differ significantly with respect
to morphology, inflection, word order, etc.
We have also studied the evolution of the similarity error rate when scaling up
to 1.4 million sentences, drawn from an out-of-domain corpus.

\section*{Acknowledgments}

We would like to thank
Ke Tran (\mbox{Informatics} Institute University of Amsterdam, \textrm{m.k.tran@uva.nl})
and Orhan Firat (Middle East Technical University, \textrm{orhan.firat@ceng.metu.edu.tr}, now at Google)
for their help with implementing
some of the algorithms during their internship at Facebook AI Research in 2016.

\newcommand{\LB}{\linebreak[4]}
\newcommand{\EX}[1]{\vspace{8pt}\underline{\bf #1}\\[6pt]}


\begin{table*}[t]
\small
\begin{tabular}[t]{|lp{0.38\textwidth}|lp{0.30\textwidth}|}
\hline
\MC{4}{|c|}{} \\[-8pt]
\bf Query: & \bf All kinds of obstacles must be eliminated.
    	& \bf Query: & \bf I did not find out why. \\[1pt]	
\hline
\MC{4}{|c|}{} \\[-8pt]
$D_2$=0.9051 & All kinds of barriers have to be removed.
	& $D_2$=0.8360 & I do not understand why. \\
$D_3$=0.6829 & All forms of violence must be prohibited.
	& $D_3$=0.8213 & I fail to understand why. \\
$D_4$=0.6738 & All forms of provocation must be avoided.
	& $D_4$=0.7862 & I cannot understand why. \\
$D_5$=0.6367 & All forms of social dumping must be stopped.
	& $D_5$=0.7804 & I have no idea why. \\
\hline
\end{tabular}
\normalsize
\caption{Five closest sentences found by monolingual similarity search in English.
 	They are some form of para-phrasing as long as the cosine distance is clsoe enough to 1.0.
	The closest sentence (distance=1) is always identical to the query and therefore omitted.}
\label{Ex1a}
\end{table*}

\begin{table*}
\small
\begin{tabular}[t]{|lp{0.85\textwidth}|}
\hline
\MC{2}{|c|}{} \\[-8pt]
\bf Query & \bf All citizens who commit sexual crimes against children must be punished, regardless of whether the crime is committed within or outside the EU. \\[1pt]
\hline
\MC{2}{|c|}{} \\[-8pt]
$D_2$=0.6626 & The second proposal is to protect children against child sex tourism by all member states criminalising sexual crimes both within and outside the EU. \\
$D_3$=0.6553 & We need standard national legislation throughout Europe which punishes union citizens who engage in child sex tourism, irrespective of where the offence was committed. \\
$D_4$=0.6553 & The impunity of those who commit terrible crimes against their own citizens and against other people regardless of their citizenship must be ended. \\
$D_5$=0.6099 & Any person who commits a criminal act should be punished, including those who employ the third-country nationals, illegally and under poor conditions. \\
\hline
\end{tabular}
\normalsize
\caption{A more complicated English sentence and the five closest sentences (excluding itself).
	All cover the punishment of (sexual) crimes.}
\label{Ex1b}
\end{table*}

\begin{table*}
  \small
  \begin{tabular}[t]{|llp{0.75\textwidth}|}

\hline
\MC{3}{|c|}{} \\[-8pt]
\bf EN$_{59177}$ & \bf Query & \bf Allow me, however, to comment on certain issues raised by the honourable Members. \\[1pt]
\hline
\MC{3}{|c|}{} \\[-8pt]
FR$_{59177}$  & $D_1$=0.7397 & Permettez-moi toutefois de commenter certaines questions soulevées par les députés. \\
FR$_{394434}$ & $D_2$=0.6435 & Je voudrais commenter quelques-unes des questions soulevées par les députés. \\
FR$_{791798}$ & $D_3$=0.6180 & Je voudrais faire les commentaires suivants sur plusieurs aspects spécifiques soulevés par certains orateurs. \\
FR$_{666349}$ & $D_4$=0.6155 & Permettez-moi de dire quelques mots sur certaines questions qui ont été soulevées. \\
FR$_{444790}$ & $D_5$=0.6090 & Je voudrais juste faire quelques commentaires sur certaines des questions qui ont été soulevées. \\
\hline
%
\MC{3}{|c|}{} \\[-8pt]
ES$_{59177}$  & $D_1$=0.7193 & No obstante, permítanme comentar ciertas cuestiones planteadas por sus señorías. \\
ES$_{394434}$ & $D_2$=0.6280 & Me gustaría comentar algunas de las cuestiones planteadas por algunos diputados. \\
ES$_{271614}$ & $D_3$=0.6155 & No obstante, quisiera hacer algunos comentarios sobre el debate que nos ocupa. \\
ES$_{661451}$ & $D_4$=0.6058 & Por ultimo, permítanme que añada algunos comentarios sobre las enmiendas presentadas. \\
ES$_{666285}$ & $D_5$=0.6055 & No obstante, permítanme que conteste a algunos comentarios que se han realizado. \\
\hline
  \end{tabular}
  \normalsize
  \caption{\textbf{Cross-lingual similarity search}. English query and the five closest French and Spanish sentences.
	We also provide the index of the sentences (reference=59177).
	All the cosine distances are close and the sentences are indeed semantically related.
  }
  \label{Ex2a}
\end{table*}

\begin{table*}
  \small
  \begin{tabular}[t]{|llp{0.75\textwidth}|}

\hline
\MC{3}{|c|}{} \\[-8pt]
\bf EN$_{77622}$ & \bf Query & \bf And yet the report on the fight against racism does not demonstrate that the necessary conclusions have been drawn. \\[1pt]
\hline
\MC{3}{|c|}{} \\[-8pt]
FR$_{77622}$   & $D_1$=0.7672 & Pourtant, le rapport sur la lutte contre le racisme n'indique pas que l'on en ait tiré les conclusions qui s'imposent. \\
FR$_{1094939}$ & $D_2$=0.7468 & Ainsi, le rapport sur la lutte contre le racisme n'indique pas que l'on en a tiré les conclusions qui s'imposent. \\
FR$_{73928}$   & $D_3$=0.4918 & Et, comme le démontrent les faits, ce n'est pas en interdisant que l'on va obtenir des résultats. \\
FR$_{1249269}$ & $D_4$=0.4761 & Ce rapport, qui se propose de lutter contre la corruption, ne fait qu'illustrer votre incapacité à le faire. \\
\hline
%
\MC{3}{|c|}{} \\[-8pt]
ES$_{77622}$   & $D_1$=0.8200 & Sin embargo, el informe sobre la lucha contra el racismo no muestra que se hayan extraído las conclusiones necesarias. \\
ES$_{1094939}$ & $D_2$=0.7973 & Así, el informe sobre la lucha contra el racismo no muestra que se hayan extraído las conclusiones necesarias. \\
ES$_{287052}$  & $D_3$=0.5172 & No obstante, el informe deja mucho que desear en lo que se refiere a las medidas necesarias para combatir el cambio climático y, por tanto, pone de relieve que el parlamento europeo no se encuentra a la vanguardia de esta batalla. \\
ES$_{74892}$   & $D_4$=0.5150 & Y el informe de los expertos demuestra que no había el control y el seguimiento necesarios. \\
\hline
  \end{tabular}
  \normalsize
  \caption{\textbf{Cross-lingual similarity search.} English query and the four closest French and Spanish sentences.
	In both cases, the correct translation was retrieved.
	The second closest sentences are also semantically well related to the query.
   	However, the third (and following sentences) only cover some of the aspects of the query.
	This is indeed reflected in the lower similarity score.
  }
  \label{Ex2b}
\end{table*}

\clearpage

\bibliographystyle{acl_natbib}
\balance
\bibliography{strings_short,repl4nlp}

\end{document}